\documentclass[a4paper]{cas-dc}

\usepackage[authoryear,longnamesfirst]{natbib}
\usepackage{booktabs}
\usepackage{tabularx}
\usepackage{amsmath,amssymb,amsfonts}
\DeclareMathOperator*{\argmax}{arg\max}
\usepackage{algorithm}
\usepackage{algpseudocode}  
\usepackage{amsmath}
\usepackage{bm}

\def\tsc#1{\csdef{#1}{\textsc{\lowercase{#1}}\xspace}}
\tsc{WGM}
\tsc{QE}
\begin{document}
\let\WriteBookmarks\relax
\let\printorcid\relax
\def\floatpagepagefraction{1}
\def\textpagefraction{.001}

% Short title
\shorttitle{An Advantage-based Optimization Method for Reinforcement Learning in Large Action Space}    

% Short author
\shortauthors{Hai Lin, Cheng Huang, and Zhihong Chen}  

% Main title of the paper
\title [mode = title]{An Advantage-based Optimization Method for Reinforcement Learning in Large Action Space}  

% Title footnote mark
% eg: \tnotemark[1]

% Title footnote 1.
% eg: \tnotetext[1]{Title footnote text}

% First author
%
% Options: Use if required
% eg: \author[1,3]{Author Name}[type=editor,
%       style=chinese,
%       auid=000,
%       bioid=1,
%       prefix=Sir,
%       orcid=0000-0000-0000-0000,
%       facebook=<facebook id>,
%       twitter=<twitter id>,
%       linkedin=<linkedin id>,
%       gplus=<gplus id>]

\author[1, 2]{Hai Lin}
\author[1, 2]{Cheng Huang}
\cormark[1]
\ead{huangchengb@whu.edu.cn}
\author[1, 2]{Zhihong Chen}

% Address/affiliation
\affiliation[1]{organization={Key Laboratory of Aerospace Information Security and Trusted Computing, Ministry of Education},
            city={Wuhan},
%          citysep={}, % Uncomment if no comma needed between city and postcode
            postcode={430072}, 
            state={Hubei},
            country={China}}

\affiliation[2]{organization={School of Cyber Science and Engineering, Wuhan University},
            city={Wuhan},
%          citysep={}, % Uncomment if no comma needed between city and postcode
            postcode={430072}, 
            state={Hubei},
            country={China}}

% Corresponding author text
\cortext[1]{Corresponding author}

% Footnote text
\fntext[1]{This work was supported in part by Hubei Province Key Research and Development Program (2023BAB022), and in part by Hubei Province International Science and Technology Collaboration Program (2023EHA033)}

% For a title note without a number/mark
%\nonumnote{}

% Here goes the abstract
\begin{abstract}
Reinforcement learning tasks in real-world scenarios often involve large, high-dimensional action spaces, leading to challenges such as convergence difficulties, instability, and high computational complexity. It is widely acknowledged that traditional value-based reinforcement learning algorithms struggle to address these issues effectively. A prevalent approach involves generating independent sub-actions within each dimension of the action space. However, this method introduces bias, hindering the learning of optimal policies. In this paper, we propose an advantage-based optimization method and an algorithm named Advantage Branching Dueling Q-network (ABQ). ABQ incorporates a baseline mechanism to tune the action value of each dimension, leveraging the advantage relationship across different sub-actions. With this approach, the learned policy can be optimized for each dimension. Empirical results demonstrate that ABQ outperforms BDQ, achieving 3\%, 171\%, and 84\% more cumulative rewards in \textit{HalfCheetah}, \textit{Ant}, and \textit{Humanoid} environments, respectively. Furthermore, ABQ exhibits competitive performance when compared against two continuous action benchmark algorithms, DDPG and TD3.
\end{abstract}

% Use if graphical abstract is present
%\begin{graphicalabstract}
%\includegraphics{}
%\end{graphicalabstract}

% Research highlights
%\begin{highlights}
%\item 
%\item 
%\item 
%\end{highlights}

%\nocite{*}

% Keywords
% Each keyword is seperated by \sep

\maketitle

% Numbered list
% Use the style of numbering in square brackets.
% If nothing is used, default style will be taken.
%\begin{enumerate}[a)]
%\item 
%\item 
%\item 
%\end{enumerate}  

% Unnumbered list
%\begin{itemize}
%\item 
%\item 
%\item 
%\end{itemize}  

% Description list
%\begin{description}
%\item[]
%\item[] 
%\item[] 
%\end{description}  

% Uncomment and use as the case may be
%\begin{theorem} 
%\end{theorem}

% Uncomment and use as the case may be
%\begin{lemma} 
%\end{lemma}

%% The Appendices part is started with the command \appendix;
%% appendix sections are then done as normal sections
%% \appendix

\section{Introduction}\label{sec1}
%% 介绍强化学习
With the wide applications in poker games \citep{wang2017research}, robot control \citep{zhang2018robot}, autonomous driving \citep{zhang2019self, huang2019end}, traffic control \citep{yang2014applications, yang2018multi}, etc., Deep Reinforcement learning (DRL) has shown great potential and has received more and more attention. In terms of implementation, DRL usually uses a neural network to approximate the value function (value-based method) or to represent the probability distribution (policy-based method) \citep{chen2021survey}. Despite the neural network's efficacy in handling high-dimensional state spaces, it faces challenges of high-dimensional action spaces. This increase in dimensionality results in an exponentially growing number of actions, often referred to as the \textit{curse of dimensionality}. 

When a task involves multiple discrete actions on each of the multiple dimensions, the issues become ever more severe. The policy-based algorithms can only generate an approximate policy, sampled from a continuous probability distribution for each action dimension. Accumulating throughout dimensions, the bias between policy and executing action will be tricky. Moreover, the policy-based algorithms have problems of convergence, sample efficiency, and training variance. On the other side, the value-based algorithms are primarily designed for discrete action spaces, which maintain value functions for each discrete action. They are deemed to adapt to discrete control tasks better and some renowned works, including Q-Learning \citep{watkins1989learning}, DQN (Deep Q-Network) \citep{mnih2013playing}, Double DQN \citep{van2016deep}, Dueling Architecture \citep{wang2016dueling}, among others, are widely adopted. However, the value-based algorithms are characterized to explicitly evaluate all possible actions, while the \textit{curse of dimensionality} contributes to overly large action spaces and innumerable actions to evaluate. They suffer from poor learning efficiency in this scenario. Thus, it's significant to enhance the training efficiency of the value-based algorithms.

Previous works address the challenge of the large action space by reducing the scale of the action space, such as action space split \citep{tavakoli2018action}, value function segmentation \citep{sunehag2017value} \citep{ rashid2020monotonic}, or action subsets processing \citep{liang2023splitnet}. By partitioning the task into sub-parts with smaller action spaces, these methods aim to alleviate computational complexity. However, after solving these sub-parts, integrating the sub-part solutions back into the original action space brings a big challenge. Many existing approaches simply concatenate all results, potentially leading to significant bias. To address this challenge, we propose an advantage-based optimization method. The advantage of an action represents the difference in Q value between that action and others. In our proposition, it is measured by a baseline which is calculated by considering the action values across all subparts. Subsequently, the action values of all subparts are tuned according to the baseline, so as to enhance the overall performance. 

%% contribution
The main contributions of this paper are summarized as follows: 
\begin{itemize}
\item We propose a DRL architecture called Advantage Branching Architecture to address the large action space issue. In this architecture, a baseline is defined based on the maximum average of all sub-parts' action values and is then used to tune these sub-actions. The tuned action values of each sub-part can reflect their advantage relationship. Through this method, the bias introduced in integration can be largely alleviated.
\item Based on Advantage Branching Architecture, we propose a novel reinforcement learning algorithm, the Advantage Branching Dueling Q-network (ABQ). By partitioning the action space into branches, ABQ significantly reduces the scale of evaluation. After that, ABQ employs the proposed baseline mechanism to tune the evaluations, enabling each branch to learn an effective policy by affecting the loss function. 
\end{itemize}

%% 章节安排
The remainder of this paper is organized as follows. Section \ref{sec2} provides an overview of previous works and their approaches to solving large action space problems. Section \ref{sec3} presents our proposed advantage-based optimization method and the implementation of the ABQ algorithm. Section \ref{sec4} evaluates the performance of the proposed algorithm against state-of-the-art algorithms. Finally, we conclude our work in Section \ref{sec5}.

%% 相关工作
\section{Related works}\label{sec2}
%% 直接分解的方法
Given that a large action space often manifests as high-dimensional, a straightforward approach involves dividing the action space along each dimension into multiple sub-action spaces and solving each sub-action space independently. This can be regarded as a Multi-Agent Reinforcement Learning (MARL) issue. \citet{tampuu2017multiagent} devised independent Q networks for the two players (two agents) in the Pong game. These agents could engage in competitive or cooperative gameplay by tuning the reward function. However, this design violates the Markovian environment requirement. Under this scheme, one player perceives the other player as part of the environment, leading to the next state being determined not only by the current state but also by the other player's actions \citep{sun2020overview}. While effective in specific experiments, this method encounters challenges when applied to more complex tasks.

%% 分解值函数
Some recent works decompose the action space indirectly through the decomposition of the value function. By keeping correlations between actions, these methods ensure the algorithm complexity increases linearly with the action dimensions. \citet{sunehag2017value} introduced Value-Decomposition Networks (VDN), which linearly decomposes the reward and the Q function, allowing for relatively independent learning of policies by multiple agents. \citet{rashid2020monotonic} identified VDN's inability to handle complex value functions and proposed a solution called QMIX. QMIX employs a nonlinear hybrid network to decompose the value function and guarantees consistency between the centralized and decentralized policies by constraining the weights of each network to be non-negative. The MARL Q-value Decomposition algorithm (MAQD) introduced by \citet{zou2021maqd} decomposes the Q value output by the critic network in the Actor-Critic (AC) architecture. It leverages the Nash equilibrium to ensure that each agent can receive the highest reward. Similarly, \citet{ming2023cooperative} proposed the CMRL (Cooperative Modular RL) method, which decomposes the large discrete action space into multiple sub-action sets. Each critic learns a decomposed value function in parallel, facilitating cooperative learning.

%% 动作分解
\citet{tavakoli2018action} introduced the Action Branching Architecture, which divides the action space into branches along dimensions and implements it as the Branching Dueling Q-network (BDQ) algorithm. To fulfill the Markovian environment requirement, they incorporated a shared feature extraction module to ensure that all branches receive the same observation. \citet{metz2017discrete} employed a next-step prediction strategy in discrete space to model Q-values and policies over continuous space. This approach sequentially predicts all dimensions, approximating the optimal action. \citet{andriotis2019managing} distributed the entire action space across different subsystems. Each subsystem has a limited number of actions and operates independently, thus the total number of actions is significantly reduced.

Methods proposed by the aforementioned works have found widespread application across diverse scenarios. \citet{chen2022cooperative} applied the value decomposition method to video encoding, decomposing video attributes (such as clarity and quantity) into sub-actions to cope with the large action space caused by the large number of videos and users. Similarly, \citet{tang2022leveraging} employed a linear value decomposition scheme in the medical system context for the large combinatorial action space. In addressing the 3D bin packing problem,  \citet{zhao2022learning} utilized RL for policy learning in each dimension to solve the large action space resulting from the high-resolution spatial discretization. Based on the BDQ algorithm proposed by \citet{tavakoli2018action}, \citet{choi2023deep} and \citet{penney2022prompt} decomposed large action spaces into multiple branches to tackle dynamic resource allocation problems in 5G communication systems. In addition, BDQ has been applied in various other domains, including network slicing reconfiguration \citep{wei2020network}, smart building environment control systems \citep{ding2019octopus}, industrial system maintenance \citep{bi2023condition}, and vehicle collaborative perception \citep{abdel2021vehicular}.

Several studies have developed strategies to restrict the selection range of actions, in order to reduce the action space. \citet{liang2023splitnet} developed SplitNet, a model designed to tackle the multiple traveling salesman problem. By utilizing a heuristic function, SplitNet focuses on specific segments of the action space, significantly improving training speed and generalization ability. In the actor-critic architecture proposed by \citet{iqbal2019actor}, an attention mechanism is designed for dynamic changes in the critic's focus. Consequently, the critic can select actions from only a few actors, thereby enhancing training efficiency. The improved Proximal Policy Optimization (PPO) algorithm in \citep{lu2022centralized} also uses a neural network based on the attention mechanism in the actor-critic architecture. Through the Bahdanau Attention layer, the critic can obtain the importance of all actors, which allows the critic to focus on existing actors, thereby ignoring part of the state or action space. A similar design can be found as well in the MAXQ algorithm proposed by  \citet{dietterich2000hierarchical}.

Hierarchical Reinforcement Learning (HRL) offers a systematic approach to decomposing complex tasks into manageable subtasks across different levels \citep{pateria2021hierarchical}. This methodology is believed to enhance decision-making capabilities and has been widely adopted in recent research endeavors. \citet{wu2021multi} introduced the concepts of managers and workers for robot control in complex environments. Workers are tasked with directly controlling the actions of the robot, while the manager is responsible for task assignment and coordination. \citet{zhang2022adjacency} extended hierarchical reinforcement learning by introducing an adjacency constraint, ensuring that agents located close to each other share the same goal. This constraint enhances the robustness and learning efficiency of the system.

For an industry system with many components, \citet{zhou2022maintenance} proposed a hierarchical coordinated RL algorithm. They introduced a hierarchical structure where components with low priority consist of subsystems. Since these subsystems are assigned with abstract actions, the overall action space is largely simplified. Abstract actions are also designed in the recommendation system domain by \citet{liu2023exploration}. Initially, the authors use a network to learn a potential feature vector, referred to as the hyper-action. Subsequently, they utilize a kernel function to transform this hyper-action into the actual action, known as the effect-action. This approach alleviates the need for the agent to directly handle the large action space, thereby enhancing algorithm efficiency.

Some other studies have focused on policy or action approximation techniques. \citet{zhang2018fully} proposed two decentralized actor-critic architectures grounded in the new policy gradient theory. In these architectures, actors share estimated action values with neighboring actors, forming a common local estimate. By voting for only a few representative actions, the number of actions that are required to be evaluated by the critic is significantly reduced. \citet{khan2018scalable} explored the scenarios involving a large number of identical agents. They indicate that agents within a certain range exhibit similar strategies that can be approximated by a common strategy. This method doesn't learn for every single agent, thus reducing the computational overhead and resource demands. For information exchange problems among homogeneous robots, they introduced the Graph Convolutional Neural (GCN) network based on Convolutional Neural Networks (CNNs)\citep{khan2020graph}. GCN aggregates information from neighboring nodes (e.g., based on Euclidean distance), while CNN's pooling operation reduces the search dimension and constrains the search space. \citet{li2022using} proposed the concept of a fuzzy agent instead of multiple agents. 

Numerous methods have been developed to tackle tasks involving large discrete action spaces. In these works, the decomposition method based on the Action Branching Architecture has emerged as a widely adopted approach due to its robust generative capabilities, versatility across tasks, and efficient training performance. Extensive research demonstrates the efficiency of the Action Branching Architecture in resolving reinforcement learning tasks with large discrete action spaces for both simulated and real-world environments. However, a notable limitation of this architecture lies in its trivial strategy for action selection, wherein each branch independently chooses its sub-actions, and the final action is just a concatenation of these sub-actions. It is obvious that the concatenation of optimal sub-actions does not always yield an optimal global action. In this paper, we introduce a novel advantage-based optimization method based on the original Action Branching Architecture. By enhancing the coordination of policies among branches, our proposed method aims to address the shortcomings of the existing architecture.

\section{Advantage Branching Dueling Q-Network}\label{sec3}

\subsection{Action Branching Architecture}\label{s3sub1}
To demonstrate the curse of dimension problem in large action space, we take an example as follows: consider a scenario with 2 dimensions, each offering 25 possible actions. The combined number of actions is ${25}^2=625$. With 3 dimensions, this number increases dramatically to ${25}^3=15625$. Formally, for an action space $\mathcal{A}$ with $n$ dimensions and $N$ possible actions in each dimension, the size of the action space, denoted as $\left|\mathcal{A}\right|$, is calculated as:
\begin{equation}
    \left|\mathcal{A}\right|=N^n.
\end{equation}

A real-world RL task is often characterized by a high-dimensional action space with a large $n$. When each dimension requires fine division, $N$ naturally increases, leading to an exponentially larger action space. As aforementioned, value-based algorithms like DQN require the evaluation of every possible action. The issue of a large action space makes it difficult to satisfy this requirement, as it demands significant computing resources and extensive training time. Consequently, it poses a challenge for value-based algorithms to learn optimal policies.

\begin{figure}
    \centering
    \includegraphics[width=0.5\textwidth]{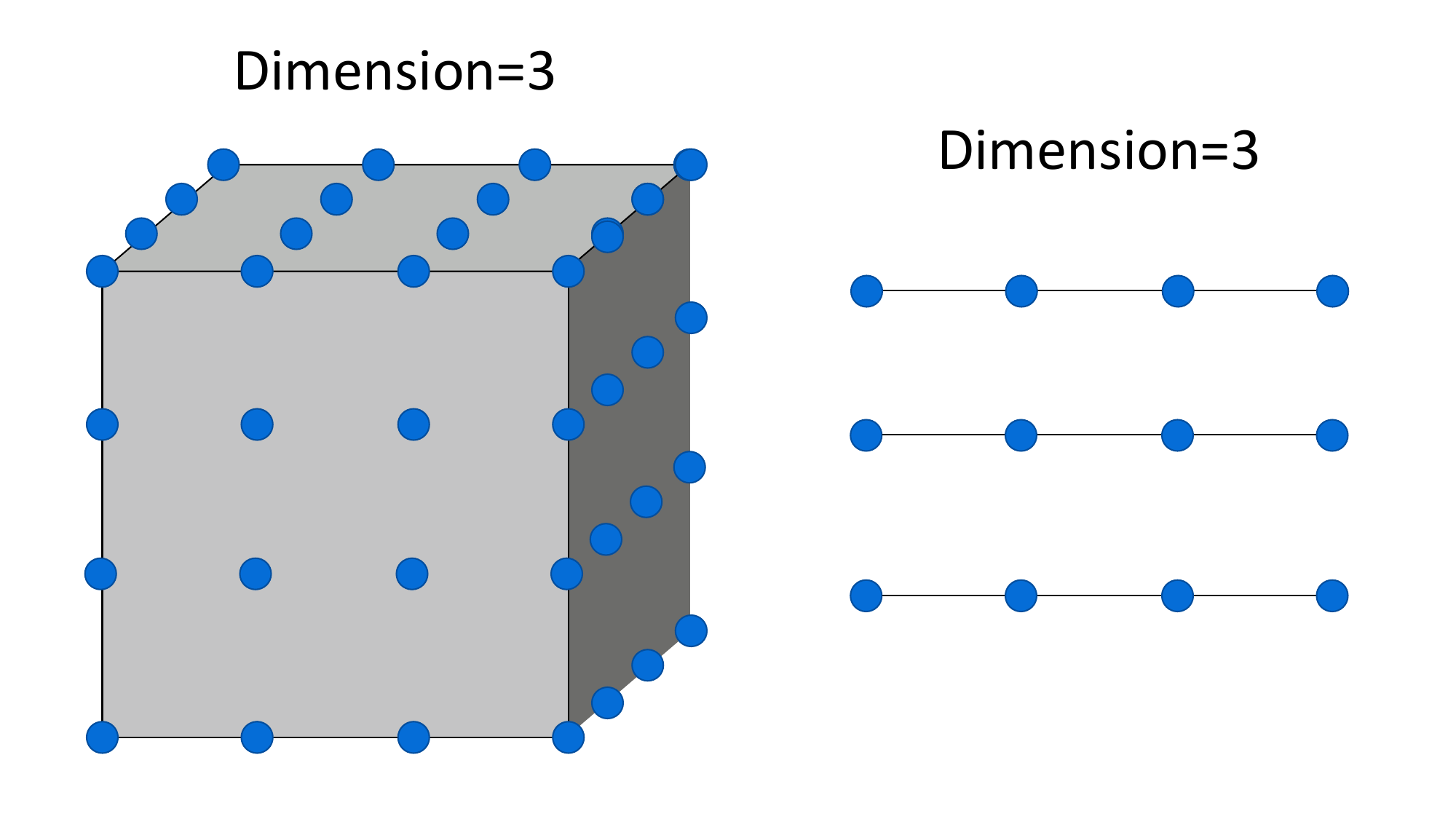}
    \caption{An example of Action Branching Architecture}
    \label{fig:branch}
\end{figure}

The Action Branching Architecture \cite{tavakoli2018action}, provides a solution for handling tasks with high-dimensional action spaces within the realm of value-based RL algorithms. This architecture partitions the action space into action branches according to dimensions, enabling the evaluation of sub-actions in each dimension. Each branch network is associated with an action dimension that assesses all sub-actions of this dimension and generates action values. The highest-valued sub-actions from each branch are then selected and combined to form a final global action. Formally, let the action $act$ comprise $n$ dimensions, denoted as $act=(act_1,\ act_2,\ldots,\ act_n)$, where $act_i$ (with $1\leq i\leq n$) represents the sub-action in the $i$th dimension. If each dimension contains $N$ possible sub-actions, then only $n\times N$ actions need to be evaluated, maintaining a linear relationship with the dimension. We take an example in Figure \ref{fig:branch}, where the action space has 3 dimensions and each contains 4 possible sub-actions. Despite the potential total of $3^4=64$ actions, only $4 \times 3 = 12$ evaluations are required. This significantly enhances decision-making efficiency.

\subsection{Advantage Branching Architecture}\label{s3sub2}
However, the combination of individual optimal sub-actions can not guarantee an optimal global action. A possible solution is to introduce interaction among branches. According to this concept, we propose an advantage-based optimization method, named the Advantage Branching Architecture. This method tunes the action value on each branch based on a pre-defined baseline. In our proposition, we update the Q value using the Temporal Difference (TD) method. For the $i$th branch, at time $t$, state $s^t$, taking action $a_i^t$, the update rule for the corresponding Q value is defined as:

\begin{equation}
\begin{aligned}
    & Q_i(s^t,a_i^t)\gets Q_i(s^t,a_i^t) + \\
    & \alpha[r^t+\gamma Q_i(s^{t+1}, a_i^{t+1}) - Q_i(s^t,a_i^t)],
\end{aligned}
\end{equation}
where $\alpha$ denotes the learning rate, $r^t$ represents the reward, $\gamma$ is the discount rate, $s^{t+1}$ and $a_i^{t+1}$ denote the subsequent state and action. The temporal difference error (TD error) is:
\begin{equation}\label{without_base}
    TDe_i=r^t+\gamma Q_i(s^{t+1},a_i^{t+1})-Q_i(s^t,a_i^t).
\end{equation}

Here, $TDe_i$ is influenced by both the reward $r^t$ and the Q value difference $\gamma Q_i(s^{t+1},a_i^{t+1}) - Q_i(s^t,a_i^t)$. As the environment determines the reward and $\gamma$ remains constant during training, value-based algorithms predominantly impact $TDe_i$ through the second term. In a multi-branch architecture, each branch may generate a distinct Q value estimate for the same state, leading to a distinct Q value difference. A branch with little Q value improvement, while others exhibiting significantly higher values, shouldn't be overly credited for its action selection. Thus a way to distinguish the advantage relationship among branches and to tune the estimate is crucial. Although the branch might rectify its estimate by increasing the Q values of other actions, employing suitable adjustments may accelerate the learning process.

Similar motivation can be found in the well-known Dueling Architecture proposed in \cite{wang2016dueling}. Unlike traditional deep reinforcement learning methods, which directly calculate the action value $Q$ through a network, the Dueling Architecture decomposes the $Q$ value into two components: the state value $V$ and the action advantage $A$:

\begin{equation}\label{dueling}
    Q(s, a)=V(s)+A(s, a),
\end{equation}
where $s$ denotes the state, $a$ represents the action, $V$ is the state value, and $A$ denotes the action advantage of action $a$ in state $s$. The core concept underlying this architecture is that in superior states, regardless of the chosen action, the $Q$ values tend to be high. By removing the effect of the state, an agent can more effectively discern the advantage of one action over the others. Inspired by this method, we introduce a baseline mechanism that discerns the advantage of one branch over the others. In our proposition, the chosen baseline, denoted as $B$, is as follows:

\begin{equation} \label{baseline}
    B = \max_{1\le i\le n}{\frac{1}{N}}\sum_{j=1}^{N}{A_i(s,\ a_{ij})},
\end{equation}
where $n$ represents the number of branches, $N$ is the number of optional sub-actions on each branch, and $a_{ij}$ denotes the $i$th optional sub-action on the $j$th branch. As we apply Dueling Architecture in our mechanism, the input is the action advantage instead of the Q value. The calculation of $B$ is critical in our proposition, defined as the maximum average action advantage among all branches. According to our experiments, this definition outperforms other baselines, such as the global maximum action advantage, the global average action advantage, etc. We decompose the action advantage on the $i$th branch further into two components: branch advantage and baseline. Then, by combining branch advantage and state value, the Q value is calculated as:

\begin{equation}\label{basic_eq}
    Q_i(s,a_{ij})=V+A_i(s,a_{ij})-B,
\end{equation}
where $A_i(s,a_{ij})-B$ represents the branch advantage of the $i$th branch over other branches. This way, the baseline is introduced to tune the action advantage. Consequently, the TD error is:

\begin{equation}\label{with_base}
\begin{aligned}
    TDe_i= r^t+\gamma &Q_i(s^{t+1},a_i^{t+1})-Q_i(s^t,a_i^t) \\
         =r^t+\gamma &(V^t+A_i(s^{t+1},a_i^{t+1})-B^t)-\\
      &(V^{t+1}+A_i(s^t,a_i^t)-B^{t+1}).
\end{aligned}
\end{equation}

By comparing equation (\ref{with_base}) to equation (\ref{without_base}), we observe that $TDe_i$ is now determined mainly by the advantage of the $i$th branch, rather than the raw output of the $i$th branch. Based on the Action Branching Architecture, we propose the Advantage Branching Architecture, which employs the baseline mechanism in equation (\ref{baseline}) and tunes the output with equation (\ref{basic_eq}). 

\begin{figure*} [hbp]
    \centering
    \includegraphics[width=1\textwidth, trim=0cm 2cm 0cm 2cm, clip]{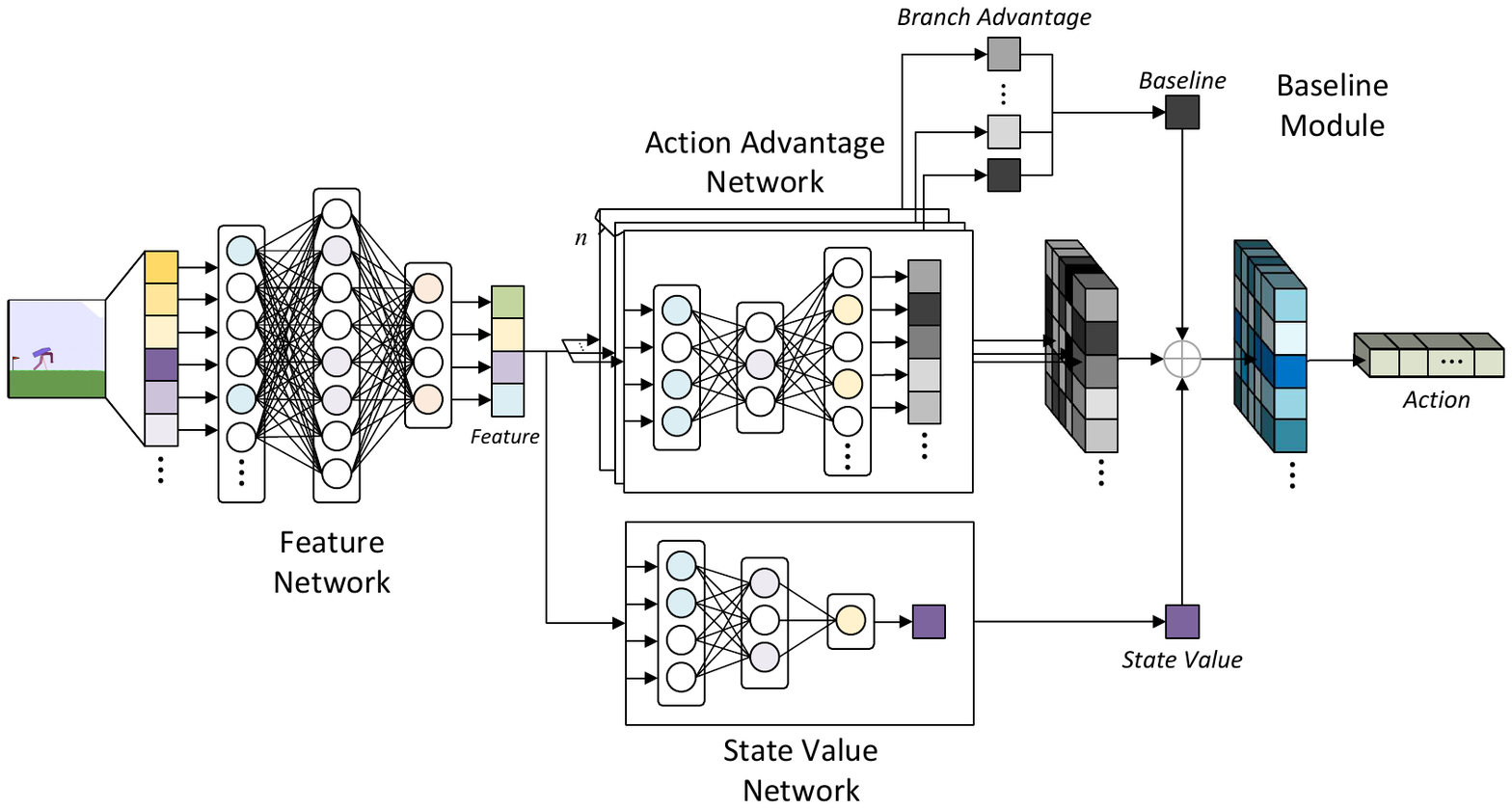}
    \caption{Advantage Branching Dueling Q-Network}
    \label{fig:network}
\end{figure*}

\subsection{Network and Algorithm}\label{s3sub3}
The network of our architecture is illustrated in Figure \ref{fig:network}, which consists of four parts:

\begin{itemize}
\item Feature Network: A shared neural network is utilized to extract features from the current state, converting the variable environment state into a fixed-length feature vector. To balance efficiency and information extraction, the neural network size is set to 256 and 128.

\item State Value Network: This part receives the state features and provides an evaluation value of the given state, as the first term in equation (\ref{dueling}). It is an essential part of the Dueling Architecture. The size of the neural network is set to 64.

\item Action Advantage Network: This network learns from the state features to generate the action advantage values as the second term in equation (\ref{dueling}). Similar to the Action Branching Architecture, it employs independent neural networks, known as branch networks, for every action dimension. Each branch evaluates the advantage values of all sub-actions in its dimensions. The size of the neural networks is set to 64.

\item Baseline Module: The Baseline Module calculates a baseline using equation (\ref{baseline}), which is then used to tune the action advantages on each branching using equation (\ref{basic_eq}). Subsequently, the tuned action advantages, together with the state value, are used to output the Q values. Sub-actions are then selected according to the maximum Q value, forming the optimal action.
\end{itemize}

The State Value Network and the Action Advantage Network form a Dueling Architecture, while the Baseline Module implements our proposed method. Our network considers the advantages of sub-actions within a branch, as well as those among different branches. Then, we develop the Advantage Branching Dueling Q-network (ABQ), where each branch network is implemented using DQN. For all sub-actions, the Q values are calculated as follows:
\begin{equation} \label{adjust}
    {Q}(s, a)= V(s) +({A}(s, a)- B).
\end{equation}

The process of ABQ is illustrated in Algorithm \ref{algo}. At the beginning, we initialize a replay buffer (line \ref{l1}), a dueling network (line \ref{l2}), and a target network (line \ref{l3}). For each iteration, we initialize the environment (line \ref{l7}) and obtain its state (line \ref{l8}). We introduce $\epsilon-greedy$ to balance exploration and exploitation (line \ref{l10}). For exploration, the agent chooses a random action. Otherwise, every branch evaluates all sub-actions in the current state, and the State Value Network generates a state value (line \ref{l13}). Then, a baseline is computed based on the output of all branches according to equation (\ref{baseline}) (line \ref{l14}), which is then applied to adjust the values of sub-actions according to equation (\ref{adjust}) (line \ref{l15}). The agent selects sub-actions with the highest values (line \ref{l16}), and executes the action in the environment (line \ref{l18}). The environment transfers to the next state based on the action and its transition probability, rewarding the agent accordingly (line \ref{l18}). Subsequently, the agent stores the tuple comprising state, reward, action, and next state in the replay buffer (line \ref{l20}). Once a sufficient number of tuples are stored in the buffer, the agent retrieves a batch of tuples (line \ref{l22}), calculates the Loss according to equation (\ref{with_base}) (line \ref{l23}), and updates its network (or policy) (Line \ref{l24}). Similar to DQN, the target network is updated every $T$ episodes (line \ref{l28}).

\begin{algorithm}[!ht]  
    \caption{Advantage Branching Dueling Q-Network}   \label{algo}
    \begin{algorithmic}[1]  
    \State Initialize replay buffer $\mathcal{D}$; \label{l1}
    \State Initialize dueling network $\mathcal{Q}$ with random weights $\theta$; \label{l2}
    \State Initialize target network $\mathcal{Q}^-$ with identical structure and weights $\theta^- \leftarrow \theta$; \label{l3} 
    \State Determine an environment $E$;
    \For{$episode \leftarrow 1, M$}
        \State $t \leftarrow 0$;
        \State $E.initialize()$;  \label{l7}
        \State $s_0 \leftarrow E.state()$; \label{l8}
        \While{not $E.is\_terminated()$} \label{l9} 
            \If{$random() < \epsilon$} \label{l10}
                \State $a_t \leftarrow random\_action()$;  
            \Else
                \State $A, V \leftarrow \mathcal{Q}(s_t)$; \label{l13}
                \State $B \leftarrow baseline(A)$ (according to Eq. (\ref{baseline})); \label{l14}
                \State $Q \leftarrow tune(A, B, V)$ (according to Eq. (\ref{adjust})); \label{l15} 
                \State $a_t \leftarrow \argmax_a(Q)$; \label{l16} 
            \EndIf
            \State $r_t \leftarrow E.execute(a_t)$; \label{l18}
            
            \State $s_{t+1} \leftarrow E.state()$; 
            \State $\mathcal{D}.store(s_{t}, a_{t}, r_{t}, s_{t+1})$; \label{l20} 
            \If{$\mathcal{D}.size() > threshold$}
                \State $(\bm{s},\bm{a},\bm{r},\bm{s'}) \leftarrow \mathcal{D}.sample(batchsize)$; \label{l22} 
            
                \State $Loss \leftarrow TDe(\bm{s},\bm{a},\bm{r},\bm{s'})$ (according to Eq. (\ref{with_base})); \label{l23}
                \State Perform a gradient descent step on $Loss$ with respect to the network parameters $\theta$; \label{l24} 
            \EndIf
            \State $t \leftarrow t + 1$;
        \EndWhile
        \State Every $T$ episodes set $\mathcal{Q}^-\leftarrow \mathcal{Q}$; \label{l28}
    \EndFor
    \end{algorithmic}
    \label{algo_flow}
\end{algorithm}

\section{Performance evaluation}\label{sec4}

Released by OpenAI on April 27, 2016, Gym provides a standardized platform for various RL tasks, including classic control problems, Atari games, 2D and 3D robot control simulations, etc. Most RL-related studies utilize Gym to train and evaluate their algorithms. For our experiments, we selected one environment with one-dimensional action space (\textit{Pendulum}) and four environments with multi-dimensional action space (\textit{BipedalWalker}, \textit{HalfCheetah}, \textit{Ant}, and \textit{Humanoid}) from Gym to analyze the performance of ABQ. In \textit{Pendulum}, the goal is to apply torque on an inverted pendulum to swing it into an upright position. In \textit{BipedalWalker}, the agent controls a 4-joint walker robot to cover a slightly uneven terrain. In \textit{HalfCheetah}, \textit{Ant}, and \textit{Humanoid}, an agent must control walker robots to move forward and avoid falling, with the control complexity increasing respectively.

Table \ref{envs} provides a summary of these environments, where the state dimension, denoted as $n_\mathcal{S}$, and the action dimension, denoted as $n_\mathcal{A}$. The complexity of the listed environments roughly increases. To accommodate value-based algorithms, we discretized each action dimension into 25 discrete values. In consequence, the simplest multi-dimensional environment \textit{BipedalWalker} with 4 action dimensions, has approximately ${25}^{4}\approx3.9e4$ discrete actions, while the most complex continuous environment \textit{Humanoid} with 17 action dimensions, yields more than ${25}^{17}\approx5.8e23$ discrete actions.

\begin{table}[width=.9\linewidth,cols=4,pos=ht]
  \caption{Information of environments}\label{envs}
  \begin{tabular*}{\tblwidth}{@{} LLLL@{} }
    \toprule
    Environment & $n_\mathcal{S}$ & $n_\mathcal{A}$ & $|\mathcal{A}|$ \\
    \midrule
    Pendulum      & 3   & 1  & 25      \\
    BipedalWalker & 24  & 4  & 3.9e4   \\
    HalfCheetah   & 17  & 6  & 2.4e8   \\
    Ant           & 27  & 8  & 1.5e11  \\
    Humanoid      & 376 & 17 & 5.8e23  \\
    \bottomrule
  \end{tabular*}
\end{table}

Besides the original BDQ, we also compare our ABQ with two benchmark continuous control algorithms: Deep Deterministic Policy Gradient (DDPG) \cite{lillicrap2015continuous} and Twin Delayed DDPG (TD3) \cite{fujimoto2018addressing}. DDPG based on actor-critic architecture employs two neural networks to approximate the policy (actor network) and the action value function (critic network) simultaneously. TD3, an extension of DDPG, enhances the stability and robustness of the training process. The implementation of ABQ and BDQ shares identical network structures and hyperparameters, while DDPG and TD3 have the similar size of the neural networks. During experiments, the hyperparameters are not fine-tuned, since we are also interested in the generalization ability of each algorithm. Each algorithm interacts with the chosen environments for 2000 episodes. The cumulative rewards for each episode are depicted using lighter colors in the figures, providing insights into the algorithm's performance. As RL algorithms typically involve extensive exploration during training, the cumulative rewards may exhibit significant fluctuations. To eliminate the fluctuation, we also apply a moving average with a window size of 100 to each reward curve, depicted with darker colors in the figures. 

\begin{figure}
    \centering
    \includegraphics[width=0.45\textwidth]{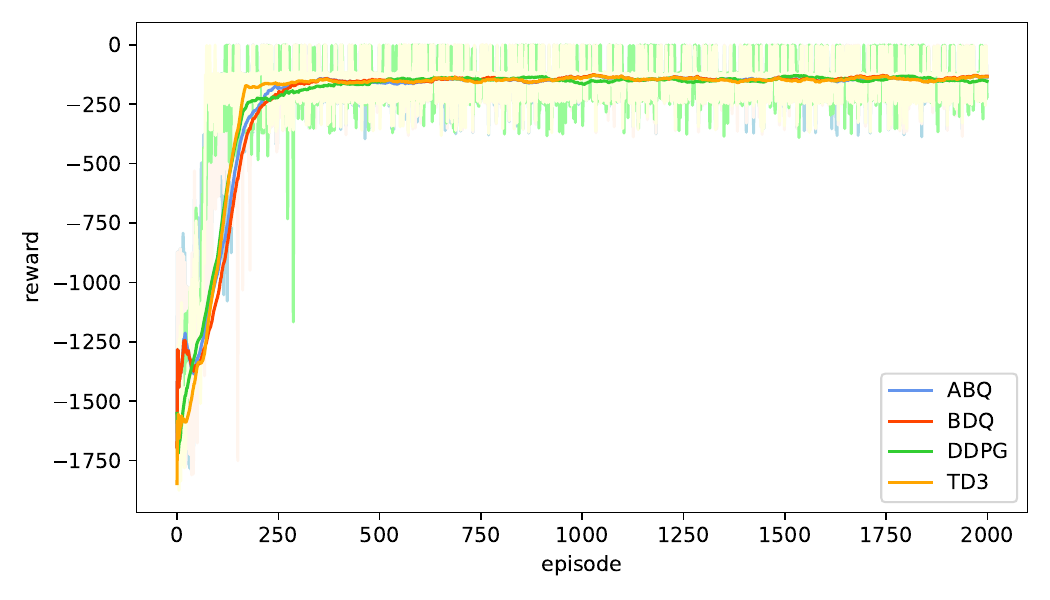}
    \caption{Performance analysis in the environment of Pendulum}
    \label{fig:pendulum}
\end{figure}

\begin{figure}
    \centering
    \includegraphics[width=0.45\textwidth]{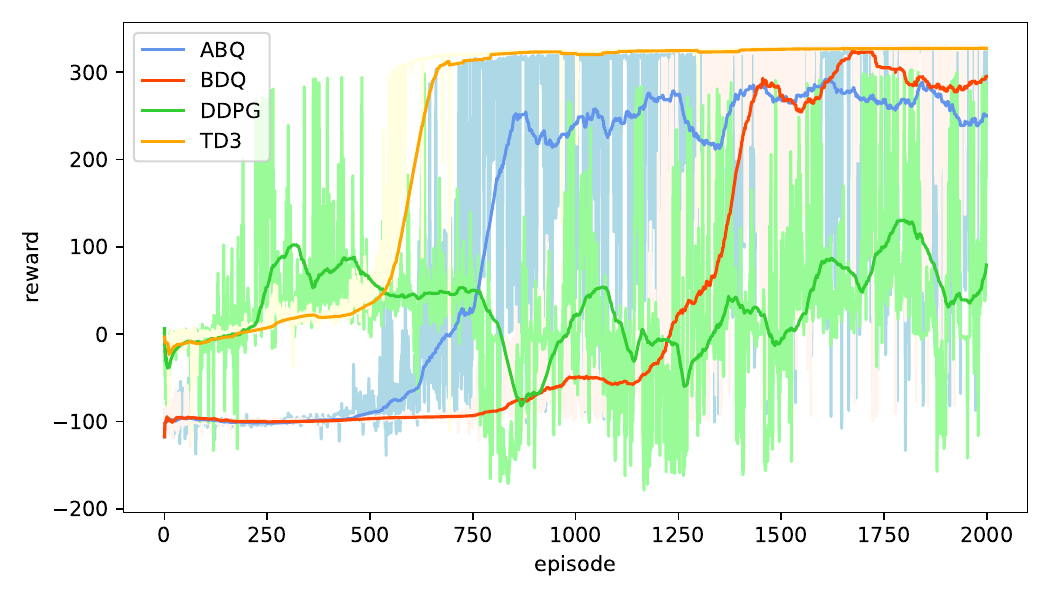}
    \caption{Performance analysis in the environment of BipedalWalker}
    \label{fig:bipedalwalker}
\end{figure}

\begin{figure}
    \centering
    \includegraphics[width=0.45\textwidth]{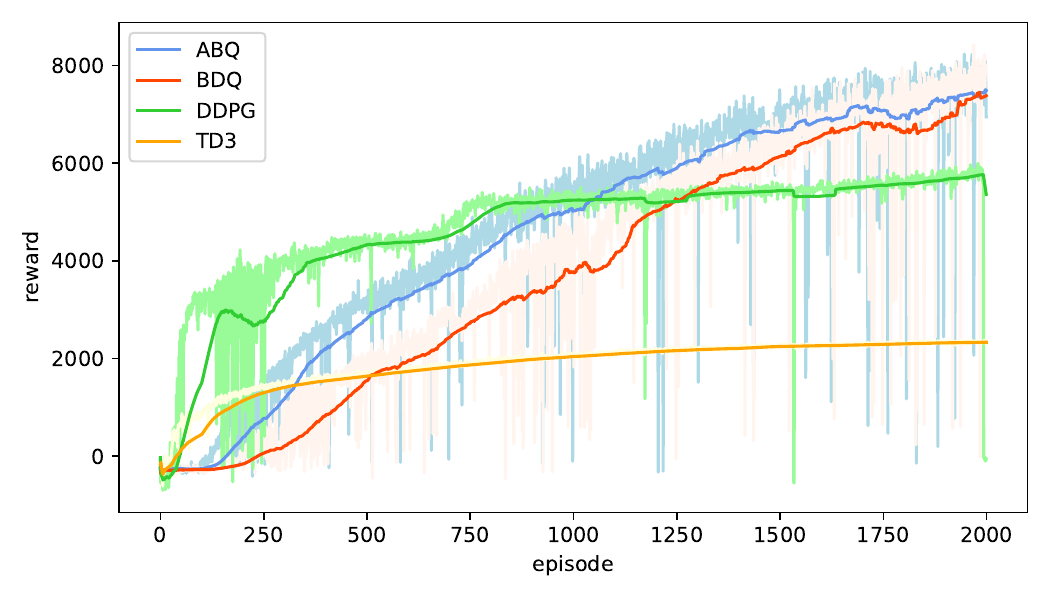}
    \caption{Performance analysis in the environment of HalfCheetah}
    \label{fig:halfcheetah}
\end{figure}

\begin{figure}
    \centering
    \includegraphics[width=0.45\textwidth]{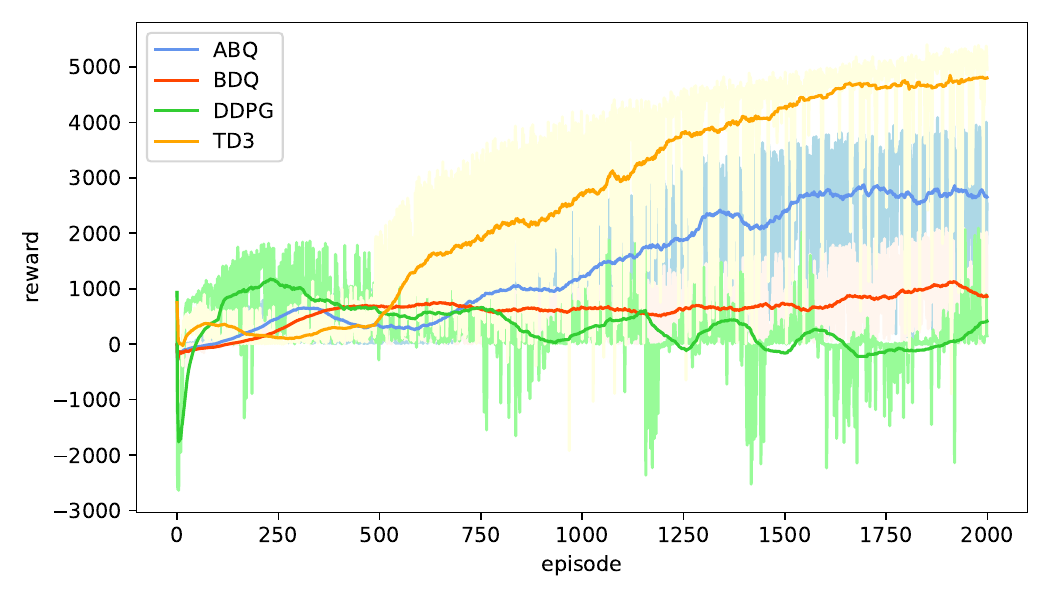}
    \caption{Performance analysis in the environment of Ant}
    \label{fig:ant}
\end{figure}

\begin{figure}
    \centering
    \includegraphics[width=0.45\textwidth]{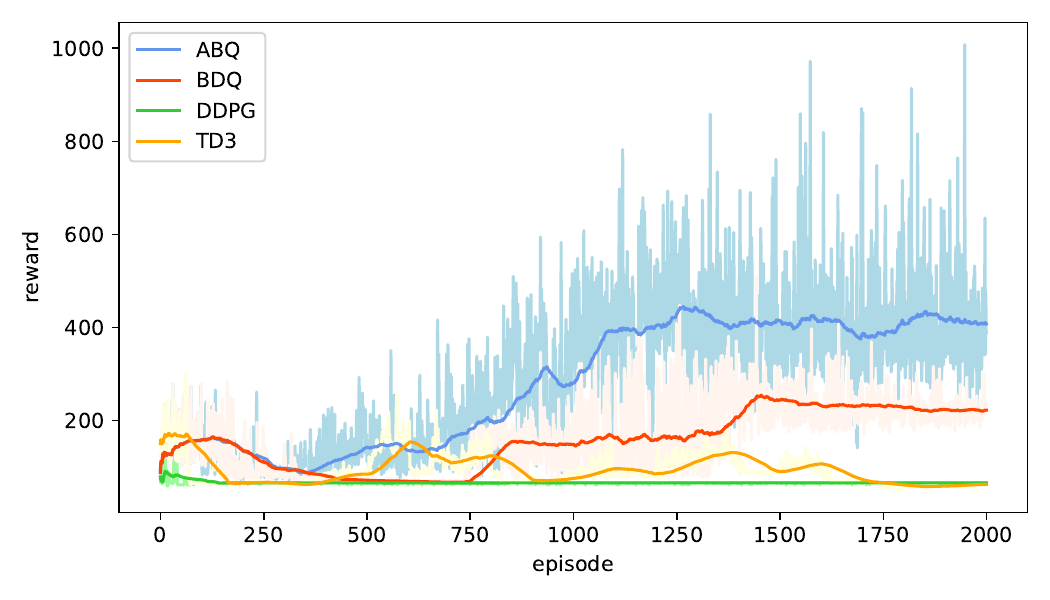}
    \caption{Performance analysis in the environment of Humanoid}
    \label{fig:humanoid}
\end{figure}

In the \textit{Pendulum} environment, the closer the pendulum is to the upright position, the less torque the agent applies to it, and the higher the reward returned. The reward in this scenario is defined with the highest value being zero. Due to its one-dimensional action space, all algorithms demonstrate similar performance, as illustrated in Figure \ref{fig:pendulum}. We observe that every algorithm converges within the first 250 episodes, with the learning curves almost overlapping. This analysis indicates that all algorithms perform well in the environment with a small action space.

The \textit{BipedalWalker} environment presents a more complex scenario. The reward is given for moving forward, totaling more than 300 points up to the far end. If the robot falls, the reward will be -100 and the episode terminates. Any torque applied to control the robot brings a negative reward. The comparison results are illustrated in Figure \ref{fig:bipedalwalker}, where we observe that DDPG performs the worst, but TD3 performs the best. This is because, although these two algorithms are similar, they are very sensitive to hyperparameter settings. It can be inferred from the results that the hyperparameters of DDPG are not well adapted to the \textit{BipedalWalker} environment, whereas those of TD3 are appropriately set. As for ABQ and BDQ, they can also bring good rewards, even though they require more training duration. Compared to BDQ, our proposition ABQ converges much faster. This is because the introduced baseline mechanism tunes the evaluation of actions properly, enhancing learning efficiency.

The comparison results in \textit{HalfCheetah} environment is shown in Figure \ref{fig:halfcheetah}. In this environment, the reward is calculated based on robots' moving distance. Similar to the \textit{BipedalWalker} environment, DDPG and TD3 converge more quickly than ABQ and BDQ. The reason stems from the fact that DDPG and TD3 directly generate actions from a continuous probability distribution, which can approximate the actual distribution after a relatively small number of episodes, while ABQ and BDQ require more training to converge, due to their discretized characteristics. An abnormal phenomenon we observe in this scenario is that TD3 performs the worst, despite having shown the best performance in our previous analysis. The reason is similar. TD3 is very sensitive to the hyperparameters setting. That is, TD3's hyperparameter setting is suitable for \textit{BipedalWalker} environment but failed in this environment. We also observe that, at convergence, the algorithms ABQ and BDQ bring the highest rewards. This demonstrates their ability to solve the task with a large action space thanks to their action branch architecture which significantly reduces the action search space. Compared to BDQ, ABQ brings higher learning efficiency due to our proposed baseline mechanism.

Figure \ref{fig:ant} illustrates the results in the \textit{Ant} environment, where the reward is defined as identical to \textit{HalfCheetah}. We observe that DDPG brings nearly zero reward. This bad performance is again due to its hyperparameter setting not fitting well to \textit{Ant} environment. In contrast, DDPG's analogue algorithm TD3 performs the best. This demonstrates both algorithms have poor generalization ability. Namely, we can not have one setting of hyperparameters for different kinds of scenarios. Therefore, these algorithms should be adaptively tuned when applied in practical scenarios. By comparing it to the \textit{HalfCheetah} scenario, we observe that our proposition ABQ outperforms BDQ even more in this scenario. This demonstrates that the baseline mechanism becomes more efficient as the action space increases.

The \textit{Humanoid} environment, with the highest dimensional state and action spaces, represents the most intricate scenario in our evaluation. The reward is defined similarly to \textit{HalfCheetah} and \textit{Ant}. The comparison results are shown in Figure \ref{fig:humanoid}. We observe that, due to large action apace, the two continuous control algorithms, DDPG and TD3, become almost like random algorithms. They fail to find any optimal policy. On the contrary, The two discrete control algorithms, BDQ and ABQ, outperform the two continuous ones. The reason is that ABQ and BDQ, leveraging the idea of action branching, are capable of handling such complexities. Furthermore, ABQ's baseline mechanism improves learning efficiency compared to BDQ, ultimately resulting in superior performance.

\begin{table}[width=.9\linewidth,cols=5,pos=ht]
  \caption{test result}\label{test}
    \begin{tabular*}{\tblwidth}{@{} LLLLL@{} }
    \toprule
    Environment & ABQ & BDQ & DDPG & TD3 \\
    \midrule
    HalfCheetah & \textbf{7476.495} & \underline{7225.685} & 5479.486 & 2327.695 \\
    Ant & \underline{2616.607} & 964.881 & 169.724 & \textbf{4702.530} \\
    Humanoid & \textbf{415.080} & \underline{225.523} & 66.003 & 61.178 \\
    \bottomrule
  \end{tabular*}
\end{table}

In the following, we conduct another 100 episodes of testing in multi-dimensional environments (i.e. \textit{HalfCheetah}, \textit{Ant}, and \textit{Humanoid}) using the policy learned through training, to further analyze the performance. Different from the training process, random exploration, and network updates are no longer used during this testing. The results are summarized in Table \ref{test}. Note that the data marked in bold indicates the best-performing algorithm, while data marked with underlines indicates the second-best-performing algorithm. We observe that ABQ achieves almost the best. ABQ outperforms all others in both \textit{HalfCheetah} and \textit{Humanoid} environments and behaves the second best in \textit{Ant}, only worse than TD3. Notably, ABQ consistently outperforms BDQ across all three environments, with improvements of 3\% in \textit{HalfCheetah}, 171\% in \textit{Ant}, and 84\% in \textit{Humanoid}. This performance demonstrates ABQ's effectiveness in addressing RL tasks with large action spaces thanks to the proposed baseline mechanism.

In summary, all algorithms, no matter whether discrete or continuous, work well in small action spaces (e.g., \textit{Pendulum} environment with only one action dimension). When the action dimension begins to increase, the two continuous action benchmark algorithms perform slightly better than the discrete action algorithms. This is simply because our scenario has continuous action space and the discrete action algorithms have to discretize the continuous action space, leading to an inevitable expansion of the action space and a decrease in accuracy. Nevertheless, as the action dimension continues to increase, especially in tasks with extremely large action spaces (e.g., \textit{Humanoid} environment with 17 action dimensions), two continuous control algorithms begin to fail. They behave like random algorithms. However, ABQ’s and BDQ's capability to deal with high-dimensional action spaces compensates for the inaccuracies arising from discretization. Compared with BDQ, ABQ not only improves the learning speed but brings high rewards as well, highlighting the effectiveness of the proposed baseline mechanism. Unlike DDPG and TD3, which have generalization issues, ABQ is capable of adapting to diverse tasks.

\section{Conclusion}\label{sec5}
This paper proposes an advantage-based action branching architecture to address the issue of large action spaces in reinforcement learning (RL). By using the baseline to calculate the advantage relationships among action branches, the action values generated by each branch can be adjusted accordingly. This adjustment improves the learning rate and results in better rewards. Building upon the foundational framework of BDQ, we designed the ABQ Network and its corresponding algorithm. Our performance analysis demonstrates that the proposed ABQ outperforms BDQ in various scenarios and achieves comparable or superior performance compared to the continuous control benchmark algorithms DDPG and TD3, even in continuous action scenarios. Hence, ABQ effectively addresses the issue of large action spaces in RL. For future work, we are interested in addressing both large action space and large state space issues, and investigating how the baseline mechanism can simultaneously tackle both challenges.

% To print the credit authorship contribution details

%% Loading bibliography style file
%\bibliographystyle{model1-num-names}
\bibliographystyle{cas-model2-names}

% Loading bibliography database
\bibliography{cas-refs}

\begin{thebibliography}{44}
\expandafter\ifx\csname natexlab\endcsname\relax\def\natexlab#1{#1}\fi
\providecommand{\url}[1]{\texttt{#1}}
\providecommand{\href}[2]{#2}
\providecommand{\path}[1]{#1}
\providecommand{\DOIprefix}{doi:}
\providecommand{\ArXivprefix}{arXiv:}
\providecommand{\URLprefix}{URL: }
\providecommand{\Pubmedprefix}{pmid:}
\providecommand{\doi}[1]{\href{http://dx.doi.org/#1}{\path{#1}}}
\providecommand{\Pubmed}[1]{\href{pmid:#1}{\path{#1}}}
\providecommand{\bibinfo}[2]{#2}
\ifx\xfnm\relax \def\xfnm[#1]{\unskip,\space#1}\fi
%Type = Article
\bibitem[{Abdel-Aziz et~al.(2021)Abdel-Aziz, Perfecto, Samarakoon, Bennis and Saad}]{abdel2021vehicular}
\bibinfo{author}{Abdel-Aziz, M.K.}, \bibinfo{author}{Perfecto, C.}, \bibinfo{author}{Samarakoon, S.}, \bibinfo{author}{Bennis, M.}, \bibinfo{author}{Saad, W.}, \bibinfo{year}{2021}.
\newblock \bibinfo{title}{Vehicular cooperative perception through action branching and federated reinforcement learning}.
\newblock \bibinfo{journal}{IEEE Transactions on Communications} \bibinfo{volume}{70}, \bibinfo{pages}{891--903}.
%Type = Article
\bibitem[{Andriotis and Papakonstantinou(2019)}]{andriotis2019managing}
\bibinfo{author}{Andriotis, C.P.}, \bibinfo{author}{Papakonstantinou, K.G.}, \bibinfo{year}{2019}.
\newblock \bibinfo{title}{Managing engineering systems with large state and action spaces through deep reinforcement learning}.
\newblock \bibinfo{journal}{Reliability Engineering \& System Safety} \bibinfo{volume}{191}, \bibinfo{pages}{106483}.
%Type = Inproceedings
\bibitem[{Bi et~al.(2023)Bi, Fang, Roux and Barros}]{bi2023condition}
\bibinfo{author}{Bi, P.}, \bibinfo{author}{Fang, Y.P.}, \bibinfo{author}{Roux, M.}, \bibinfo{author}{Barros, A.}, \bibinfo{year}{2023}.
\newblock \bibinfo{title}{Condition-based maintenance optimization under large action space with deep reinforcement learning method}, in: \bibinfo{booktitle}{International Conference on Optimization and Learning}, \bibinfo{organization}{Springer}. pp. \bibinfo{pages}{161--172}.
%Type = Article
\bibitem[{Chen et~al.(2021)Chen, Yao, McAuley, Zhou and Wang}]{chen2021survey}
\bibinfo{author}{Chen, X.}, \bibinfo{author}{Yao, L.}, \bibinfo{author}{McAuley, J.}, \bibinfo{author}{Zhou, G.}, \bibinfo{author}{Wang, X.}, \bibinfo{year}{2021}.
\newblock \bibinfo{title}{A survey of deep reinforcement learning in recommender systems: A systematic review and future directions}.
\newblock \bibinfo{journal}{arXiv preprint arXiv:2109.03540} .
%Type = Article
\bibitem[{Chen et~al.(2022)Chen, Cai, Zheng, Hu and Li}]{chen2022cooperative}
\bibinfo{author}{Chen, Y.}, \bibinfo{author}{Cai, Y.}, \bibinfo{author}{Zheng, H.}, \bibinfo{author}{Hu, J.}, \bibinfo{author}{Li, J.}, \bibinfo{year}{2022}.
\newblock \bibinfo{title}{Cooperative caching for scalable video coding using value-decomposed dimensional networks}.
\newblock \bibinfo{journal}{China Communications} \bibinfo{volume}{19}, \bibinfo{pages}{146--161}.
%Type = Article
\bibitem[{Choi et~al.(2023)Choi, Choi, Lee, Yoon and Bahk}]{choi2023deep}
\bibinfo{author}{Choi, S.}, \bibinfo{author}{Choi, S.}, \bibinfo{author}{Lee, G.}, \bibinfo{author}{Yoon, S.G.}, \bibinfo{author}{Bahk, S.}, \bibinfo{year}{2023}.
\newblock \bibinfo{title}{Deep reinforcement learning for scalable dynamic bandwidth allocation in ran slicing with highly mobile users}.
\newblock \bibinfo{journal}{IEEE Transactions on Vehicular Technology} .
%Type = Article
\bibitem[{Dietterich(2000)}]{dietterich2000hierarchical}
\bibinfo{author}{Dietterich, T.G.}, \bibinfo{year}{2000}.
\newblock \bibinfo{title}{Hierarchical reinforcement learning with the maxq value function decomposition}.
\newblock \bibinfo{journal}{Journal of artificial intelligence research} \bibinfo{volume}{13}, \bibinfo{pages}{227--303}.
%Type = Inproceedings
\bibitem[{Ding et~al.(2019)Ding, Du and Cerpa}]{ding2019octopus}
\bibinfo{author}{Ding, X.}, \bibinfo{author}{Du, W.}, \bibinfo{author}{Cerpa, A.}, \bibinfo{year}{2019}.
\newblock \bibinfo{title}{Octopus: Deep reinforcement learning for holistic smart building control}, in: \bibinfo{booktitle}{Proceedings of the 6th ACM international conference on systems for energy-efficient buildings, cities, and transportation}, pp. \bibinfo{pages}{326--335}.
%Type = Inproceedings
\bibitem[{Fujimoto et~al.(2018)Fujimoto, Hoof and Meger}]{fujimoto2018addressing}
\bibinfo{author}{Fujimoto, S.}, \bibinfo{author}{Hoof, H.}, \bibinfo{author}{Meger, D.}, \bibinfo{year}{2018}.
\newblock \bibinfo{title}{Addressing function approximation error in actor-critic methods}, in: \bibinfo{booktitle}{International conference on machine learning}, \bibinfo{organization}{PMLR}. pp. \bibinfo{pages}{1587--1596}.
%Type = Inproceedings
\bibitem[{Huang et~al.(2019)Huang, Zhang, Tian and Zhang}]{huang2019end}
\bibinfo{author}{Huang, Z.}, \bibinfo{author}{Zhang, J.}, \bibinfo{author}{Tian, R.}, \bibinfo{author}{Zhang, Y.}, \bibinfo{year}{2019}.
\newblock \bibinfo{title}{End-to-end autonomous driving decision based on deep reinforcement learning}, in: \bibinfo{booktitle}{2019 5th International Conference on Control, Automation and Robotics (ICCAR)}, \bibinfo{organization}{IEEE}. pp. \bibinfo{pages}{658--662}.
%Type = Inproceedings
\bibitem[{Iqbal and Sha(2019)}]{iqbal2019actor}
\bibinfo{author}{Iqbal, S.}, \bibinfo{author}{Sha, F.}, \bibinfo{year}{2019}.
\newblock \bibinfo{title}{Actor-attention-critic for multi-agent reinforcement learning}, in: \bibinfo{booktitle}{International conference on machine learning}, \bibinfo{organization}{PMLR}. pp. \bibinfo{pages}{2961--2970}.
%Type = Inproceedings
\bibitem[{Khan et~al.(2020)Khan, Tolstaya, Ribeiro and Kumar}]{khan2020graph}
\bibinfo{author}{Khan, A.}, \bibinfo{author}{Tolstaya, E.}, \bibinfo{author}{Ribeiro, A.}, \bibinfo{author}{Kumar, V.}, \bibinfo{year}{2020}.
\newblock \bibinfo{title}{Graph policy gradients for large scale robot control}, in: \bibinfo{booktitle}{Conference on robot learning}, \bibinfo{organization}{PMLR}. pp. \bibinfo{pages}{823--834}.
%Type = Article
\bibitem[{Khan et~al.(2018)Khan, Zhang, Lee, Kumar and Ribeiro}]{khan2018scalable}
\bibinfo{author}{Khan, A.}, \bibinfo{author}{Zhang, C.}, \bibinfo{author}{Lee, D.D.}, \bibinfo{author}{Kumar, V.}, \bibinfo{author}{Ribeiro, A.}, \bibinfo{year}{2018}.
\newblock \bibinfo{title}{Scalable centralized deep multi-agent reinforcement learning via policy gradients}.
\newblock \bibinfo{journal}{arXiv preprint arXiv:1805.08776} .
%Type = Article
\bibitem[{Li et~al.(2022)Li, Shi and Hwang}]{li2022using}
\bibinfo{author}{Li, J.}, \bibinfo{author}{Shi, H.}, \bibinfo{author}{Hwang, K.S.}, \bibinfo{year}{2022}.
\newblock \bibinfo{title}{Using fuzzy logic to learn abstract policies in large-scale multiagent reinforcement learning}.
\newblock \bibinfo{journal}{IEEE Transactions on Fuzzy Systems} \bibinfo{volume}{30}, \bibinfo{pages}{5211--5224}.
%Type = Inproceedings
\bibitem[{Liang et~al.(2023)Liang, Ma, Cao, Liu, Ni, Li and Hao}]{liang2023splitnet}
\bibinfo{author}{Liang, H.}, \bibinfo{author}{Ma, Y.}, \bibinfo{author}{Cao, Z.}, \bibinfo{author}{Liu, T.}, \bibinfo{author}{Ni, F.}, \bibinfo{author}{Li, Z.}, \bibinfo{author}{Hao, J.}, \bibinfo{year}{2023}.
\newblock \bibinfo{title}{Splitnet: a reinforcement learning based sequence splitting method for the minmax multiple travelling salesman problem}, in: \bibinfo{booktitle}{Proceedings of the AAAI Conference on Artificial Intelligence}, pp. \bibinfo{pages}{8720--8727}.
%Type = Article
\bibitem[{Lillicrap et~al.(2015)Lillicrap, Hunt, Pritzel, Heess, Erez, Tassa, Silver and Wierstra}]{lillicrap2015continuous}
\bibinfo{author}{Lillicrap, T.P.}, \bibinfo{author}{Hunt, J.J.}, \bibinfo{author}{Pritzel, A.}, \bibinfo{author}{Heess, N.}, \bibinfo{author}{Erez, T.}, \bibinfo{author}{Tassa, Y.}, \bibinfo{author}{Silver, D.}, \bibinfo{author}{Wierstra, D.}, \bibinfo{year}{2015}.
\newblock \bibinfo{title}{Continuous control with deep reinforcement learning}.
\newblock \bibinfo{journal}{arXiv preprint arXiv:1509.02971} .
%Type = Inproceedings
\bibitem[{Liu et~al.(2023)Liu, Cai, Sun, Wang, Jiang, Zheng, Jiang, Gai, Zhao and Zhang}]{liu2023exploration}
\bibinfo{author}{Liu, S.}, \bibinfo{author}{Cai, Q.}, \bibinfo{author}{Sun, B.}, \bibinfo{author}{Wang, Y.}, \bibinfo{author}{Jiang, J.}, \bibinfo{author}{Zheng, D.}, \bibinfo{author}{Jiang, P.}, \bibinfo{author}{Gai, K.}, \bibinfo{author}{Zhao, X.}, \bibinfo{author}{Zhang, Y.}, \bibinfo{year}{2023}.
\newblock \bibinfo{title}{Exploration and regularization of the latent action space in recommendation}, in: \bibinfo{booktitle}{Proceedings of the ACM Web Conference 2023}, pp. \bibinfo{pages}{833--844}.
%Type = Article
\bibitem[{Lu et~al.(2022)Lu, Bao, Xia and Qu}]{lu2022centralized}
\bibinfo{author}{Lu, C.}, \bibinfo{author}{Bao, Q.}, \bibinfo{author}{Xia, S.}, \bibinfo{author}{Qu, C.}, \bibinfo{year}{2022}.
\newblock \bibinfo{title}{Centralized reinforcement learning for multi-agent cooperative environments}.
\newblock \bibinfo{journal}{Evolutionary Intelligence} , \bibinfo{pages}{1--7}.
%Type = Article
\bibitem[{Metz et~al.(2017)Metz, Ibarz, Jaitly and Davidson}]{metz2017discrete}
\bibinfo{author}{Metz, L.}, \bibinfo{author}{Ibarz, J.}, \bibinfo{author}{Jaitly, N.}, \bibinfo{author}{Davidson, J.}, \bibinfo{year}{2017}.
\newblock \bibinfo{title}{Discrete sequential prediction of continuous actions for deep rl}.
\newblock \bibinfo{journal}{arXiv preprint arXiv:1705.05035} .
%Type = Article
\bibitem[{Ming et~al.(2023)Ming, Gao, Liu and Zhao}]{ming2023cooperative}
\bibinfo{author}{Ming, F.}, \bibinfo{author}{Gao, F.}, \bibinfo{author}{Liu, K.}, \bibinfo{author}{Zhao, C.}, \bibinfo{year}{2023}.
\newblock \bibinfo{title}{Cooperative modular reinforcement learning for large discrete action space problem}.
\newblock \bibinfo{journal}{Neural Networks} \bibinfo{volume}{161}, \bibinfo{pages}{281--296}.
%Type = Article
\bibitem[{Mnih et~al.(2013)Mnih, Kavukcuoglu, Silver, Graves, Antonoglou, Wierstra and Riedmiller}]{mnih2013playing}
\bibinfo{author}{Mnih, V.}, \bibinfo{author}{Kavukcuoglu, K.}, \bibinfo{author}{Silver, D.}, \bibinfo{author}{Graves, A.}, \bibinfo{author}{Antonoglou, I.}, \bibinfo{author}{Wierstra, D.}, \bibinfo{author}{Riedmiller, M.}, \bibinfo{year}{2013}.
\newblock \bibinfo{title}{Playing atari with deep reinforcement learning}.
\newblock \bibinfo{journal}{arXiv preprint arXiv:1312.5602} .
%Type = Article
\bibitem[{Pateria et~al.(2021)Pateria, Subagdja, Tan and Quek}]{pateria2021hierarchical}
\bibinfo{author}{Pateria, S.}, \bibinfo{author}{Subagdja, B.}, \bibinfo{author}{Tan, A.h.}, \bibinfo{author}{Quek, C.}, \bibinfo{year}{2021}.
\newblock \bibinfo{title}{Hierarchical reinforcement learning: A comprehensive survey}.
\newblock \bibinfo{journal}{ACM Computing Surveys (CSUR)} \bibinfo{volume}{54}, \bibinfo{pages}{1--35}.
%Type = Article
\bibitem[{Penney et~al.(2022)Penney, Li, Sydir, Tai, Walsh, Long, Lee and Chen}]{penney2022prompt}
\bibinfo{author}{Penney, D.}, \bibinfo{author}{Li, B.}, \bibinfo{author}{Sydir, J.}, \bibinfo{author}{Tai, C.}, \bibinfo{author}{Walsh, E.}, \bibinfo{author}{Long, T.}, \bibinfo{author}{Lee, S.}, \bibinfo{author}{Chen, L.}, \bibinfo{year}{2022}.
\newblock \bibinfo{title}{Prompt: Learning dynamic resource allocation policies for edge-network applications}.
\newblock \bibinfo{journal}{arXiv preprint arxiv:2201.07916} .
%Type = Article
\bibitem[{Rashid et~al.(2020)Rashid, Samvelyan, De~Witt, Farquhar, Foerster and Whiteson}]{rashid2020monotonic}
\bibinfo{author}{Rashid, T.}, \bibinfo{author}{Samvelyan, M.}, \bibinfo{author}{De~Witt, C.S.}, \bibinfo{author}{Farquhar, G.}, \bibinfo{author}{Foerster, J.}, \bibinfo{author}{Whiteson, S.}, \bibinfo{year}{2020}.
\newblock \bibinfo{title}{Monotonic value function factorisation for deep multi-agent reinforcement learning}.
\newblock \bibinfo{journal}{The Journal of Machine Learning Research} \bibinfo{volume}{21}, \bibinfo{pages}{7234--7284}.
%Type = Article
\bibitem[{Sun et~al.(2020)Sun, Cao, Chen et~al.}]{sun2020overview}
\bibinfo{author}{Sun, Y.}, \bibinfo{author}{Cao, L.}, \bibinfo{author}{Chen, X.}, et~al., \bibinfo{year}{2020}.
\newblock \bibinfo{title}{Overview of multi-agent deep reinforcement learning}.
\newblock \bibinfo{journal}{Comput. Eng. Appl} \bibinfo{volume}{56}, \bibinfo{pages}{13--24}.
%Type = Article
\bibitem[{Sunehag et~al.(2017)Sunehag, Lever, Gruslys, Czarnecki, Zambaldi, Jaderberg, Lanctot, Sonnerat, Leibo, Tuyls et~al.}]{sunehag2017value}
\bibinfo{author}{Sunehag, P.}, \bibinfo{author}{Lever, G.}, \bibinfo{author}{Gruslys, A.}, \bibinfo{author}{Czarnecki, W.M.}, \bibinfo{author}{Zambaldi, V.}, \bibinfo{author}{Jaderberg, M.}, \bibinfo{author}{Lanctot, M.}, \bibinfo{author}{Sonnerat, N.}, \bibinfo{author}{Leibo, J.Z.}, \bibinfo{author}{Tuyls, K.}, et~al., \bibinfo{year}{2017}.
\newblock \bibinfo{title}{Value-decomposition networks for cooperative multi-agent learning}.
\newblock \bibinfo{journal}{arXiv preprint arXiv:1706.05296} .
%Type = Article
\bibitem[{Tampuu et~al.(2017)Tampuu, Matiisen, Kodelja, Kuzovkin, Korjus, Aru, Aru and Vicente}]{tampuu2017multiagent}
\bibinfo{author}{Tampuu, A.}, \bibinfo{author}{Matiisen, T.}, \bibinfo{author}{Kodelja, D.}, \bibinfo{author}{Kuzovkin, I.}, \bibinfo{author}{Korjus, K.}, \bibinfo{author}{Aru, J.}, \bibinfo{author}{Aru, J.}, \bibinfo{author}{Vicente, R.}, \bibinfo{year}{2017}.
\newblock \bibinfo{title}{Multiagent cooperation and competition with deep reinforcement learning}.
\newblock \bibinfo{journal}{PloS one} \bibinfo{volume}{12}, \bibinfo{pages}{e0172395}.
%Type = Article
\bibitem[{Tang et~al.(2022)Tang, Makar, Sjoding, Doshi-Velez and Wiens}]{tang2022leveraging}
\bibinfo{author}{Tang, S.}, \bibinfo{author}{Makar, M.}, \bibinfo{author}{Sjoding, M.}, \bibinfo{author}{Doshi-Velez, F.}, \bibinfo{author}{Wiens, J.}, \bibinfo{year}{2022}.
\newblock \bibinfo{title}{Leveraging factored action spaces for efficient offline reinforcement learning in healthcare}.
\newblock \bibinfo{journal}{Advances in Neural Information Processing Systems} \bibinfo{volume}{35}, \bibinfo{pages}{34272--34286}.
%Type = Inproceedings
\bibitem[{Tavakoli et~al.(2018)Tavakoli, Pardo and Kormushev}]{tavakoli2018action}
\bibinfo{author}{Tavakoli, A.}, \bibinfo{author}{Pardo, F.}, \bibinfo{author}{Kormushev, P.}, \bibinfo{year}{2018}.
\newblock \bibinfo{title}{Action branching architectures for deep reinforcement learning}, in: \bibinfo{booktitle}{Proceedings of the aaai conference on artificial intelligence}.
%Type = Inproceedings
\bibitem[{Van~Hasselt et~al.(2016)Van~Hasselt, Guez and Silver}]{van2016deep}
\bibinfo{author}{Van~Hasselt, H.}, \bibinfo{author}{Guez, A.}, \bibinfo{author}{Silver, D.}, \bibinfo{year}{2016}.
\newblock \bibinfo{title}{Deep reinforcement learning with double q-learning}, in: \bibinfo{booktitle}{Proceedings of the AAAI conference on artificial intelligence}.
%Type = Article
\bibitem[{Wang(2017)}]{wang2017research}
\bibinfo{author}{Wang, P.}, \bibinfo{year}{2017}.
\newblock \bibinfo{title}{Research on imperfect information machine game based on deep reinforcement learning}.
\newblock \bibinfo{journal}{Harbin Institute of Technology} .
%Type = Inproceedings
\bibitem[{Wang et~al.(2016)Wang, Schaul, Hessel, Hasselt, Lanctot and Freitas}]{wang2016dueling}
\bibinfo{author}{Wang, Z.}, \bibinfo{author}{Schaul, T.}, \bibinfo{author}{Hessel, M.}, \bibinfo{author}{Hasselt, H.}, \bibinfo{author}{Lanctot, M.}, \bibinfo{author}{Freitas, N.}, \bibinfo{year}{2016}.
\newblock \bibinfo{title}{Dueling network architectures for deep reinforcement learning}, in: \bibinfo{booktitle}{International conference on machine learning}, \bibinfo{organization}{PMLR}. pp. \bibinfo{pages}{1995--2003}.
%Type = Article
\bibitem[{Watkins(1989)}]{watkins1989learning}
\bibinfo{author}{Watkins, C.J.C.H.}, \bibinfo{year}{1989}.
\newblock \bibinfo{title}{Learning from delayed rewards} .
%Type = Article
\bibitem[{Wei et~al.(2020)Wei, Feng, Sun, Wang, Qin and Liang}]{wei2020network}
\bibinfo{author}{Wei, F.}, \bibinfo{author}{Feng, G.}, \bibinfo{author}{Sun, Y.}, \bibinfo{author}{Wang, Y.}, \bibinfo{author}{Qin, S.}, \bibinfo{author}{Liang, Y.C.}, \bibinfo{year}{2020}.
\newblock \bibinfo{title}{Network slice reconfiguration by exploiting deep reinforcement learning with large action space}.
\newblock \bibinfo{journal}{IEEE Transactions on Network and Service Management} \bibinfo{volume}{17}, \bibinfo{pages}{2197--2211}.
%Type = Article
\bibitem[{Wu et~al.(2021)Wu, Gao, Wang, Zhang and Liu}]{wu2021multi}
\bibinfo{author}{Wu, M.}, \bibinfo{author}{Gao, Y.}, \bibinfo{author}{Wang, P.}, \bibinfo{author}{Zhang, F.}, \bibinfo{author}{Liu, Z.}, \bibinfo{year}{2021}.
\newblock \bibinfo{title}{The multi-dimensional actions control approach for obstacle avoidance based on reinforcement learning}.
\newblock \bibinfo{journal}{Symmetry} \bibinfo{volume}{13}, \bibinfo{pages}{1335}.
%Type = Article
\bibitem[{Yang et~al.(2014)Yang, Zhang, Shi and Zhang}]{yang2014applications}
\bibinfo{author}{Yang, W.}, \bibinfo{author}{Zhang, L.}, \bibinfo{author}{Shi, Y.}, \bibinfo{author}{Zhang, M.}, \bibinfo{year}{2014}.
\newblock \bibinfo{title}{Applications of agent technology in urban traffic signal control systems: A survey}.
\newblock \bibinfo{journal}{Journal of Wuhai University of Technology (Transportation Science Engineering)} \bibinfo{volume}{38}, \bibinfo{pages}{709--718}.
%Type = Article
\bibitem[{Yang et~al.(2018)Yang, Zhang and Zhu}]{yang2018multi}
\bibinfo{author}{Yang, W.}, \bibinfo{author}{Zhang, L.}, \bibinfo{author}{Zhu, F.}, \bibinfo{year}{2018}.
\newblock \bibinfo{title}{Multi-agent reinforcement learning based traffic signal control for integrated urban network: Survey of state of art}.
\newblock \bibinfo{journal}{Appl. Res. Comput} \bibinfo{volume}{35}.
%Type = Article
\bibitem[{Zhang et~al.(2019)Zhang, He, Chen et~al.}]{zhang2019self}
\bibinfo{author}{Zhang, B.}, \bibinfo{author}{He, M.}, \bibinfo{author}{Chen, X.}, et~al., \bibinfo{year}{2019}.
\newblock \bibinfo{title}{Self-driving via improved ddpg algorithm}.
\newblock \bibinfo{journal}{Computer Engineering and Applications} \bibinfo{volume}{55}, \bibinfo{pages}{264--270}.
%Type = Article
\bibitem[{Zhang et~al.(2018a)Zhang, Li, Yuan and Fu}]{zhang2018robot}
\bibinfo{author}{Zhang, F.}, \bibinfo{author}{Li, N.}, \bibinfo{author}{Yuan, R.}, \bibinfo{author}{Fu, Y.}, \bibinfo{year}{2018}a.
\newblock \bibinfo{title}{Robot path planning algorithm based on reinforcement learning}.
\newblock \bibinfo{journal}{Journal of Huazhong University of Science and Technology (Natural Science Edition)} \bibinfo{volume}{46}, \bibinfo{pages}{70--75}.
%Type = Inproceedings
\bibitem[{Zhang et~al.(2018b)Zhang, Yang, Liu, Zhang and Basar}]{zhang2018fully}
\bibinfo{author}{Zhang, K.}, \bibinfo{author}{Yang, Z.}, \bibinfo{author}{Liu, H.}, \bibinfo{author}{Zhang, T.}, \bibinfo{author}{Basar, T.}, \bibinfo{year}{2018}b.
\newblock \bibinfo{title}{Fully decentralized multi-agent reinforcement learning with networked agents}, in: \bibinfo{booktitle}{International Conference on Machine Learning}, \bibinfo{organization}{PMLR}. pp. \bibinfo{pages}{5872--5881}.
%Type = Article
\bibitem[{Zhang et~al.(2022)Zhang, Guo, Tan, Hu and Chen}]{zhang2022adjacency}
\bibinfo{author}{Zhang, T.}, \bibinfo{author}{Guo, S.}, \bibinfo{author}{Tan, T.}, \bibinfo{author}{Hu, X.}, \bibinfo{author}{Chen, F.}, \bibinfo{year}{2022}.
\newblock \bibinfo{title}{Adjacency constraint for efficient hierarchical reinforcement learning}.
\newblock \bibinfo{journal}{IEEE Transactions on Pattern Analysis and Machine Intelligence} \bibinfo{volume}{45}, \bibinfo{pages}{4152--4166}.
%Type = Article
\bibitem[{Zhao et~al.(2022)Zhao, Zhu, Xu, Huang and Xu}]{zhao2022learning}
\bibinfo{author}{Zhao, H.}, \bibinfo{author}{Zhu, C.}, \bibinfo{author}{Xu, X.}, \bibinfo{author}{Huang, H.}, \bibinfo{author}{Xu, K.}, \bibinfo{year}{2022}.
\newblock \bibinfo{title}{Learning practically feasible policies for online 3d bin packing}.
\newblock \bibinfo{journal}{Science China Information Sciences} \bibinfo{volume}{65}, \bibinfo{pages}{112105}.
%Type = Article
\bibitem[{Zhou et~al.(2022)Zhou, Li and Lin}]{zhou2022maintenance}
\bibinfo{author}{Zhou, Y.}, \bibinfo{author}{Li, B.}, \bibinfo{author}{Lin, T.R.}, \bibinfo{year}{2022}.
\newblock \bibinfo{title}{Maintenance optimisation of multicomponent systems using hierarchical coordinated reinforcement learning}.
\newblock \bibinfo{journal}{Reliability Engineering \& System Safety} \bibinfo{volume}{217}, \bibinfo{pages}{108078}.
%Type = Inproceedings
\bibitem[{Zou et~al.(2021)Zou, Jiang, Li, Gao, Li and He}]{zou2021maqd}
\bibinfo{author}{Zou, Q.}, \bibinfo{author}{Jiang, Y.}, \bibinfo{author}{Li, W.}, \bibinfo{author}{Gao, B.}, \bibinfo{author}{Li, D.}, \bibinfo{author}{He, M.}, \bibinfo{year}{2021}.
\newblock \bibinfo{title}{Maqd: Cooperative multi-agent reinforcement learning q-value decomposition in actor-critic framework}, in: \bibinfo{booktitle}{2021 China Automation Congress (CAC)}, \bibinfo{organization}{IEEE}. pp. \bibinfo{pages}{8194--8199}.

\end{thebibliography}

% Biography
%\bio{}
% Here goes the biography details.
%\endbio

%\bio{pic1}
% Here goes the biography details.
%\endbio

\end{document}